\documentclass[10pt,twocolumn,letterpaper]{article}

\usepackage{iccv}
\usepackage{times}
\usepackage{epsfig}
\usepackage{graphicx}
\usepackage{amsmath}
\usepackage{amssymb}

\usepackage{bm}
\usepackage{bbm}
\usepackage{pifont}
\usepackage{tabularx}
\usepackage{subfigure}
\usepackage{amsmath}
\usepackage{amssymb}
\usepackage{multirow}
\usepackage{algorithm}
\newcommand{\xmark}{\ding{55}}%

\usepackage[breaklinks=true,bookmarks=false]{hyperref}

\iccvfinalcopy 

\def\iccvPaperID{****} 

\begin{document}

\title{Unsupervised Noisy Tracklet Person Re-identification}

\author{Minxian Li\\
Nanjing University of Science and Technology\\
{\tt\small minxianli@njust.edu.cn}
\and
Xiatian Zhu\\
Vision Semantics Limited\\
{\tt\small eddy.zhuxt@gmail.com}
\and
Shaogang Gong\\
Queen Mary University of London\\
{\tt\small s.gong@qmul.ac.uk}
}

\maketitle

\begin{abstract}
Existing person re-identification (re-id)
methods mostly rely on supervised model learning from 
a large set of person identity labelled training data per domain. 
This limits their scalability and usability
in large scale deployments.
In this work, we present a novel {\em selective tracklet learning} (STL)
approach that can train discriminative person re-id models
from unlabelled tracklet data in an {\em unsupervised} manner.
This avoids the tedious and costly process of
exhaustively labelling person image/tracklet true matching pairs
across camera views.
Importantly, our method is particularly more robust
against arbitrary noisy data of raw tracklets
therefore scalable to learning discriminative models
from unconstrained tracking data.
This differs from a handful of existing alternative methods
that often assume the existence of true 
matches and balanced tracklet samples per identity class. 
This is achieved by formulating a
data adaptive image-to-tracklet selective 
matching loss function explored 
in a multi-camera multi-task deep learning model structure.
Extensive comparative experiments demonstrate that 
the proposed STL model surpasses significantly
the state-of-the-art unsupervised learning
and one-shot learning re-id methods 
on three large tracklet person re-id benchmarks. 
\end{abstract}

\section{Introduction}

\begin{figure}[!ht] 
	\centering
	\subfigure
	{\includegraphics[width=1\linewidth]{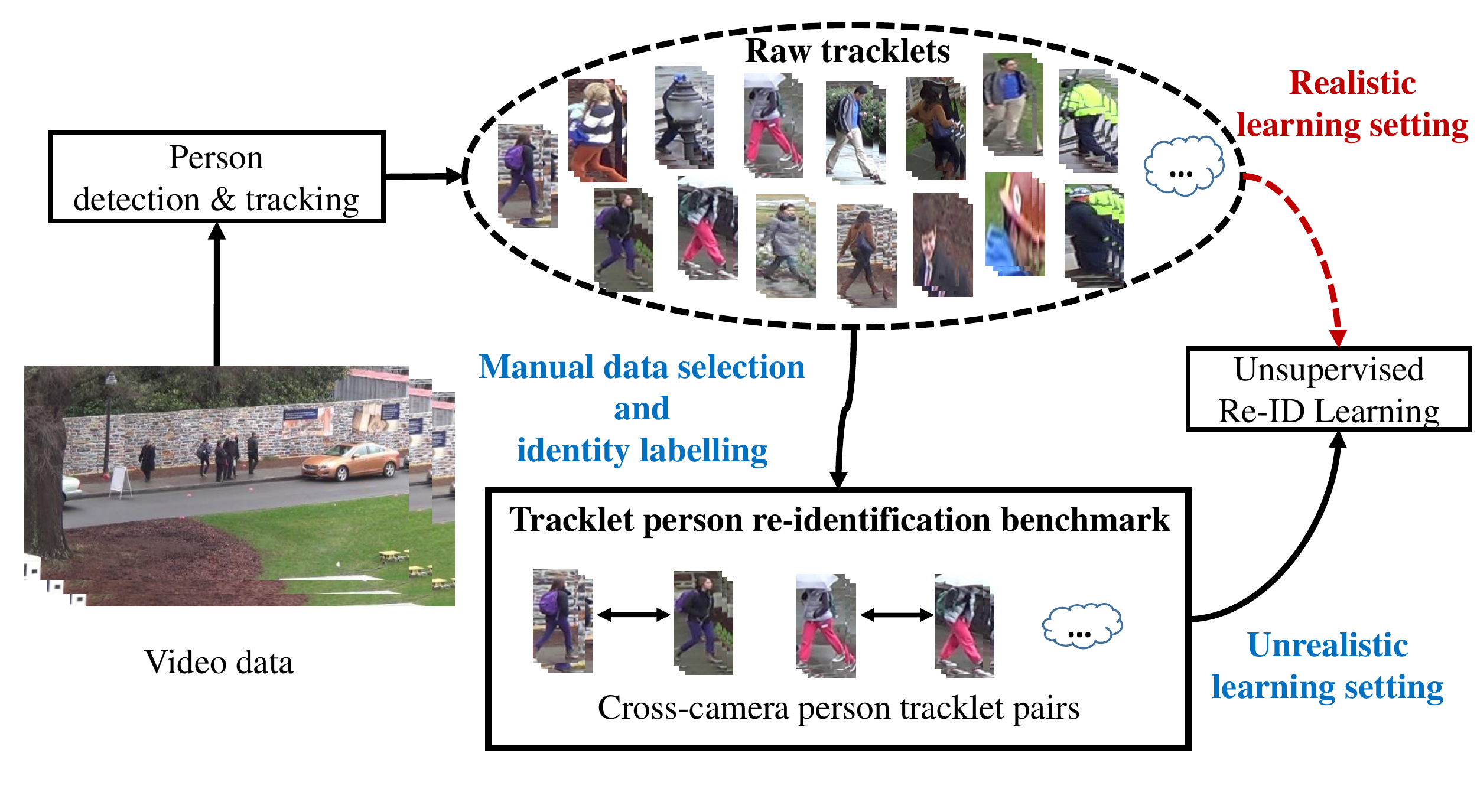}}
	\subfigure
	{\includegraphics[width=1\linewidth]{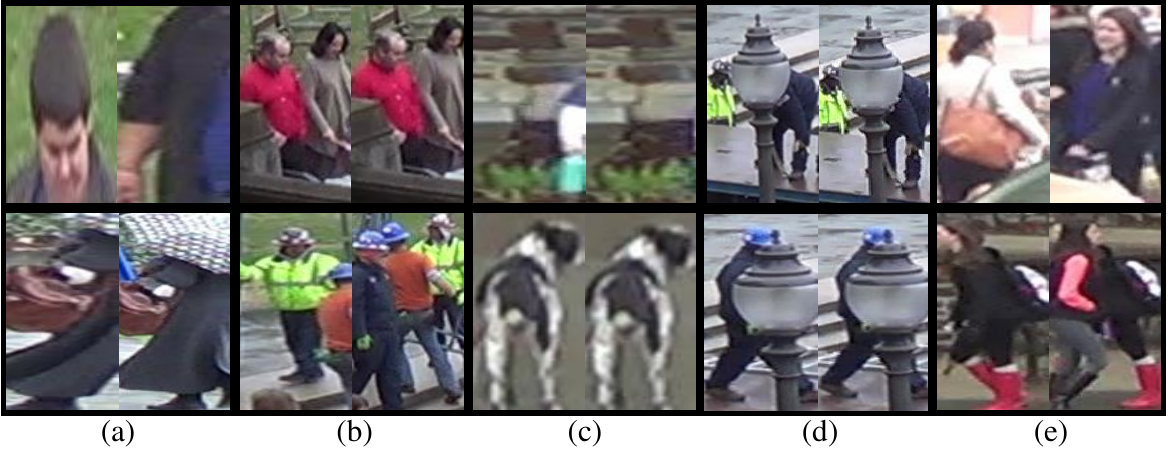}}
	\caption{{\bf (Top)} Existing tracklet person re-id benchmarks exhibit {\em less}
	realistic evaluation scenarios for unsupervised model learning.
	This is due to data selection from manual annotation
	which gives rises to an easier learning task with 
	significantly less noisy data. In current benchmarks, tracklets without
        cross-camera matches are excluded, and poor tracklets are often
        discarded including {\bf(Bottom)}:  
	(a) Partial detection, 
	(b) Multiple persons, 
	(c) Non-person,
	(d) Severe occlusion,
	(e) Identity switch.
}
\label{fig:raw_tkls}
\end{figure}

Person re-identification (re-id) is a task of matching the identity information 
of person bounding box images extracted from
disjoint surveillance camera views \cite{gong2014person}.
Existing state-of-the-art re-id methods 
rely heavily on {\em supervised deep learning} 
\cite{xiao2016learning,li2017person,li2018harmonious,wei2018person,song2018mask,chang2018multi,shen2018deep,zhang2017deep,Shen_2018_ECCV,Suh_2018_ECCV}.
They assume a large set of 
cross-camera pairwise training data exhaustively labelled 
{\em per surveillance camera network} (i.e. domain),
and are often significantly degraded for new domain deployments.
Such poor cross-domain scalability 
leads to recent research focus on 
developing unsupervised domain adaptation 
\cite{zheng2017unlabeled,deng2018image,zhong2018generalizing,peng2016unsupervised,want2018Transfer}  
and unsupervised deep learning 
\cite{li2018unsupervised,LiTPAMI2019,lin2019aBottom,chen2018deep} methods.
In general, model learning with domain adaptation is less scalable
since it assumes some common characteristics between 
the source and target domains which is not always true.

Unsupervised deep learning re-id models \cite{li2018unsupervised,LiTPAMI2019,lin2019aBottom,chen2018deep}
have started increasingly to explore unlabelled tracking data (tracklets).
This is reasonable and intuitive because most image frames of a tracklet
may share the person identity (ID), which provides
rich spatio-temporal appearance variation information.
To enable discriminative model optimisation, 
the key is to self-discover and learning reliable within-camera and cross-camera 
tracklet true matching among {\em inevitably noisy tracklet training data}.
It is non-trivial due to that
person ID labelling is unavailable 
for discriminative learning and noisy tracklet frame detection, 
as well as the majority pairs are false matches.

Existing tracklet person re-id benchmarks 
(e.g. MARS \cite{zheng2015scalable} and
DukeMTMC-SI-Tracklet \cite{ristani2016performance,LiTPAMI2019}) 
present {\em artificially simplified} evaluation scenarios for unsupervised learning.
This is because after manual selection and annotation in dataset construction, 
their tracklet data are {\em no longer realistic}
(Fig. \ref{fig:raw_tkls}).
For example, all the tracklets without cross-camera true matching
and with poor conditions 
are often removed.
Manually removing such tracklets in annotation
would significantly simplify model learning,
e.g. a 10\% rank-1 rate difference in model performance \cite{LiTPAMI2019}.
In real-world applications, tracklet manual filtering is {\em not} available.
Scalable unsupervised learning algorithms are required to handle automatically
{\em unconstrained raw tracklet data} without manual data selection. 

In this work, we consider the problem of unsupervised deep learning
on unconstrained {raw tracklet data}, which is a more realistic and scalable setting
than the existing tests \cite{zheng2015scalable,ristani2016performance,LiTPAMI2019}.
Given unfiltered and unlabelled noisy training data, 
more robust tracklet learning algorithm is required.
To this end, we present a {\em selective tracklet learning} (STL)
method.
STL is characterised by a robust image-to-tracklet selective matching loss function
that is able to selectively associate true matching tracklets
and adaptively suppress potentially noisy frame images and tracklets
in model learning. 
It does not assume the existence of true matches 
for individual tracklets within and across camera views.
%


The {\bf contributions} of this work are as follows: 
(1) We analyse the limitations of
the existing tracklet person re-id benchmarks and methods.
In particular, the current benchmarks fail to reflect the
true challenges for unsupervised model learning, due to 
the effect of data selection during manual annotation.
This phenomenon makes the developed methods
less scalable and robust to more realistic scenarios.
(2) We formulate a {\em selective tracklet learning} (STL) method
for unsupervised deep learning with superior robustness
against unconstrained tracklet data.
This is achieved by designing a
data adaptive image-to-tracklet selective matching loss function.
(3) To enable more realistic unsupervised tracklet learning test,
we introduce an unconstrained raw tracklet person re-id dataset 
DukeMTMC-Raw-Tracklet.
It is constructed based on the DukeMTMC tracking benchmark \cite{ristani2016performance}.
%
%
Extensive comparative experiments show the 
performance advantages and superior robustness 
of STL over the state-of-the-art
unsupervised and one-shot learning models on three tracklet person re-id
benchmarks:
MARS \cite{zheng2016mars},
DukeMTMC-SI-Tracklet \cite{LiTPAMI2019,ristani2016performance},
and the newly introduced DukeMTMC-Raw-Tracklet.

\section{Related Work}
\noindent{\bf Supervised person re-id.}
Most existing person re-id models are {\em supervised} learning methods
with a large set of cross-camera ID labelled training data
\cite{li2014deepreid,chen2018person,li2017person,li2018harmonious,wei2018person,song2018mask,chang2018multi,shen2018deep,Shen_2018_ECCV,Suh_2018_ECCV}.
Moreover, the training and test data are typically assumed to be 
sampled from the same surveillance camera network,
i.e. the same domain.
%
As a result, their scalability and usability 
is significantly reduced for 
large real-world applications. 
This is because no such large training sets are available
in typical test domains due to high labelling costs.

\vspace{0.1cm}
\noindent {\em Supervision reduction.}
To address the scalability and generalisation limitation of 
supervised learning re-id models,
unsupervised model learning is desired.
%
%
A trade-off can be achieved by semi-supervised learning
\cite{liu2014semi,wang2016towards}, although still need labelled data. 
Alternatively, human-in-the-loop models can reduce
the labelling effort by leveraging 
human-computer interaction \cite{wang2016human,liu2013pop}, although
the process can be overly elaborated and involved.
Unsupervised model learning is attractive without the need to collect ID labelled training data.
However, earlier attempts 
\cite{farenzena2010person,ma2017person,kodirov2015dictionary,kodirov2016person,khan2016unsupervised,ye2017dynamic,liu2017stepwise,lisanti2015person,wang2014unsupervised,zhao2013unsupervised}
have rather poor re-id performance 
due to weak hand-crafted features.

\vspace{0.1cm}
\noindent {\bf Unsupervised domain adaptation person re-id.}
Recently, unsupervised domain adaptation 
methods have gained noticeable success
\cite{want2018Transfer,fan2017unsupervised,peng2018joint,yu2017cross,deng2018image,zhong2018generalizing}.
The idea is to transfer the available person identity information 
from a labelled source domain to an unlabelled target
domain.
Existing methods can be generally divided into three groups.
The first group is {\em image synthesis}
which aims to render the source person identities
into the target domain environment 
\cite{zheng2017unlabeled,deng2018image,zhong2018generalizing,bak2018domain}.
As such, the conventional supervised learning algorithms
are enabled to train re-id models.
The second group is the conventional {\em feature alignment} scheme
\cite{peng2016unsupervised,peng2018joint,want2018Transfer}.
%
%
The third group is by {\em unsupervised clustering} which generates 
pseudo labels for supervised learning
\cite{fan2017unsupervised,yu2017cross}.
%
These methods are usually stronger than unsupervised learning methods.
However, they assume a similar imagery data distribution between the source and
target domains, which restricts their generalisation to arbitrary and unconstrained
application scenarios.

\begin{figure*}[t]
	\centering
	\includegraphics[width=\textwidth]{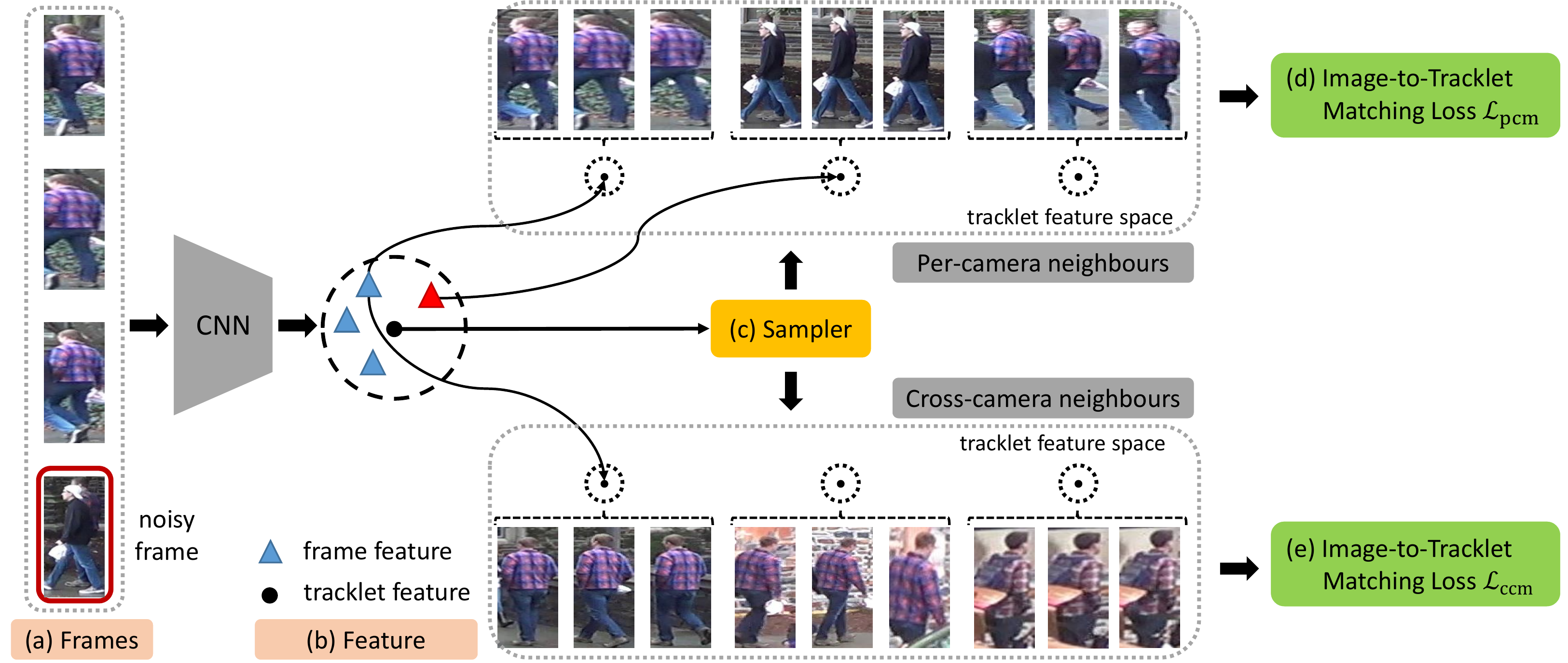}
	\caption{
		An overview of the proposed {selective tracklet learning} (STL) method
		for unsupervised tracklet re-id model learning.
		(a) The frames from the same tracklet with the noisy frame.
		(b) The frame feature and tracklet feature are both used for model learning.
		(c) An adaptive sampler is introduced to generate per-camera neighbours
		and cross-camera neighbours of tracklets.
		(d) A per-camera image-to-tracklet selective matching loss 
		$\mathcal{L}_\text{pcm}$ is proposed
		to learn the feature representation against noisy tracklet data within the same camera.
		(e) A cross-camera image-to-tracklet selective matching loss 
		$\mathcal{L}_\text{ccm}$ is proposed to learn the feature representation against noisy tracklet data across cameras.
	}
\label{fig:pipeline}
\end{figure*}

\vspace{0.1cm}
\noindent{\bf Unsupervised video tracklet person re-id.}
Unsupervised video tracklet re-id methods have been advanced
notably \cite{li2018unsupervised,chen2018deep,lin2019aBottom,LiTPAMI2019}.
They excel at leveraging the spatio-temporal continuity information
of tracklet data freely available from person tracking.
Whilst solving this tracklet learning problem is inherently challenging, 
it promises enormous potential 
due to massive surveillance videos available for model learning.
To encourage novel model development,
two large scale tracklet person re-id benchmarks
\cite{zheng2016mars,LiTPAMI2019,ristani2016performance}
were constructed.
They are based on exhaustive data selection
and identity labelling during benchmarking.
When testing unsupervised learning algorithms,
one assumes no identity labels whilst still
use the same training tracklet data.
This means the training data
are {\em selected manually}, other than the unconstrained raw tracklet data
as typically encountered in real-world unsupervised model learning.
This discrepancy on model training data renders
the existing benchmarks fail to test the 
realistic performance of unsupervised model learning algorithms.
Moreover, this data bias also gives undesired influence on
algorithm designing.
For example, current methods
\cite{li2018unsupervised,chen2018deep,LiTPAMI2019,lin2019aBottom}
often assume the existence of tracklet matches
and class balanced training data.
Such assumptions however is often not valid
on {\em truly} unconstrained tracklet data.
Consequently, they are less robust and scalable to more realistic
unfiltered tracklets (Table \ref{tab:vid_SOTA}).

In this study, we aim to resolve the aforementioned artificial and unrealistic assumption
for scaling up unsupervised learning algorithms to real-world 
application scenarios.
To this end, we propose a selective tracklet learning approach.
It is capable of learning a discriminative re-id model directly
from unlabelled raw tracklet data without manual selection and noise removal.
To enable testing true performance,
we further introduce an unconstrained raw tracklet person re-id benchmark
by using DukeMTMC videos.
%

\section{Methodology}

\noindent {\bf Problem setting.}
We start with automated person tracklet extraction on 
a large set of multi-camera surveillance videos
by off-the-shelf detection and tracking models
\cite{leal2015motchallenge,girdhar2018detect}.
Let us denote the $i$-th tracklet from the $m$-th camera 
as $\bm{T}_i^m = \{\bm{I}_1, \bm{I}_2, \cdots \}$ with
a varying number of image frames.
There is a total of $M$ camera views. 
The per-camera tracklet number $N^m$ varies.
In unsupervised learning, {\em no person ID labels} 
are available on person tracklet data.
The \textbf{\em objective} is to learn a discriminative person re-id model
from these unconstrained raw tracklets {\em without} any manual processing.

\noindent {\bf Approach overview.}
We adopt multi-task unsupervised tracklet learning
with each task dedicated for an individual camera view
as \cite{LiTPAMI2019,li2018unsupervised}.
In particular, we first {\em automatically}
annotate each tracklet with a unique class label
per-camera.
Each camera is therefore associated with an independent class label set.
%
%
In multi-task learning, we ground all the branches
on a common re-id feature representation.
%
%
This model takes individual frame images $\bm{I}$ as input
(Fig. \ref{fig:pipeline}(a)) other than tracklets. 
This is favourable due to the possibility of 
modelling noisy image frames within tracklets.
After extracting the frame feature afte the backbone CNN, 
we aggregate the frame features from the same tracklet to the tracklet feature.
Especially, we proposed an adaptive sampler to generate two neighbour sets:
{\em per-camera neighbours} and {\em cross-camera neighbours}.
Based on these two neighbour sets, 
we proposed the per-camera/cross-camera image-to-tracklet selective matching loss functions to learn the feature representation against noisy tracklet data.
An overview of our STL is depicted in Fig. \ref{fig:pipeline}.

\subsection{Per-Camera View Learning}
\label{sec:method_per_cam_learning}

For enforcing model learning constraints for each camera view, 
the softmax Cross Entropy (CE) classification loss function is utilised
\cite{LiTPAMI2019,li2018unsupervised} as:
\begin{equation}
	\mathcal{L}_\text{ce}
	={-}\sum_{i=1}^{N^m} \mathbbm{1} (i=y)
	  \cdot
	  {\log}\Big(\frac{\exp({\bm{W}_{i}^{\top} {\bm x}})}
	  {\sum_{k=1}^{N^m}\exp({\bm{W}_{k}^{\top} {\bm x}})}\Big)
\label{eq:CE_loss}
\end{equation}
where $\bm{x}$ is the {\em task-shared} feature vector of an input frame image $\bm{I}$ and
$\bm{W}_{i}$ the classifier parameters for the $i$-th tracklet label.
The indicator function $\mathbbm{1} (\cdot)$ 
returns $1$ for true arguments and $0$ for otherwise.
The term in $\log(\cdot)$ defines the posterior probability
on the tracklet label $i$.

In this context, the CE loss assumes that each tracklet is to be
matched with only a single person candidate. 
This is not valid for many raw tracklet data with 
noise introduced into model learning.
To address this limitation, we propose
a novel {\em image-to-tracklet selective matching} loss formulation.
It is a weighted non-parametric formulation of the CE loss.
Formally, the proposed per-camera image-to-tracklet selective matching loss function
for a specific training image frame $\bm{I}_i$ is designed as (Fig. \ref{fig:pipeline}(d)):
\begin{equation}
	\mathcal{L}_\text{pcm} 
	 ={-}\sum_{j=1}^{N^m}{w(i,j)}
	    \cdot
	   {\log}(P(\bm{z}_j|\bm{x}_i))
\label{eq:soft_TA_loss}
\end{equation}
where $\bm{x}_i$ and $\bm{z}_j$ 
specify the feature vector of image $\bm{I}_i$ and the $j$-th tracklet as $\bm{T}_j$,
$P(\bm{z}_j|\bm{x}_i)$ is the matching probability of
image $\bm{x}_i$ and tracklet $\bm{z}_j$,  
$N^m$ is the tracklet number of the $m$-th camera view,
$w(i,j)$ denotes the similarity weight between
the $\bm{I}_i$'s corresponding tracklet and the $j$-th tracklet $\bm{z}_j$.
This weight aims to minimise negative effect of trajectory fragmentation 
by taking into account the tracklet pairwise information
in classification \cite{LiTPAMI2019}.
The specific computation of $w(i,j)$, as defined in Eq.~(\ref{eq:matching_weight}),
will be discussed below. 

To suppress the contamination of tracklet by noisy and distracting image frames,
we introduce the posterior probability $P(\bm{z}_j|\bm{x}_i)$ based on image-to-tracklet selective matching as:
\begin{equation}
	P(\bm{z}_j|\bm{x}_i) 
	= \frac{\exp(\bm{z}^\top_j\bm{x}_i/\tau)}
	   {\sum_{k=1}^{N^m}\exp(\bm{z}^\top_k\bm{x}_i/\tau)}
\label{eq:prob}
\end{equation}
where $\bm{z}^\top_j\bm{x}_i$ expresses the matching degree
between image $\bm{x}_i$ and tracklet $\bm{z}_j$,
$\tau$ is a temperature parameter that controls the 
concentration of the distribution \cite{hinton2015distilling}. 
It is normalised over all the $N^m$ tracklets
by the softmax function.

In contrast to the point-to-point probability 
in the non-parametric classification loss \cite{wu2018unsupervised}, 
Eq. \eqref{eq:prob} is a point-to-set matching probability, 
which is more robust to contaminated and distracting tracklets.
%
This can be understood at two aspects:
(1) Suppose a tracklet contains noisy frames due to
multiple person, non-person or ID switch, etc.
Often, the noisy frames tend to have much smaller 
matching scores $\bm{z}^\top_j\bm{x}_i$ against a true-matching tracklet,
as compared to other clean frames.  
%
(2) 
The image-to-tracklet pairs with large matching scores
will become significantly more salient 
after applying the exponential operation $\exp$.
This is effectively a process of 
selecting good-quality matching tracklets (e.g. less noisy true matches) 
and simultaneously 
down-weighing the remaining ones (e.g. more noisy true matches and false matches).
If there is no true match, all $\bm{z}^\top_j\bm{x}_i$ values tend to be small.
This data adaptive and selective matching capability is highly desired for
dealing with noisy raw tracklets in unsupervised tracklet re-id learning.

%
%

In unsupervised tracklet training data, 
the majority of tracklet pairs per camera are false matches.
Therefore, considering all the pairs 
in Eq. \eqref{eq:soft_TA_loss} is likely to 
introduce a large amount of negative matching.
As \cite{LiTPAMI2019}, we consider only a fraction of tracklets
that are more likely true matches (i.e. tracklet association).
To this end, $k$ nearest neighbour ($k$-NN) search is often adopted: 
\begin{equation}
\mathcal{N}_k(\bm{z}) = \{\bm{z}' \;\; | \;\; \bm{z}'^\top\bm{z} {\text{ is among top-}k}.\} 
\label{eq:KNN}
\end{equation}
For each tracklet, this implicitly assumes $k$ true matches in each camera view.
Given unconstrained raw tracklets without manual selection, 
this condition is often harder to meet.
%
Data adaptive tracklet association is hence needed.

To that end, we suggest to further exploit the concept of $\epsilon$-neighbourhood ($\epsilon$-NN) (Fig. \ref{fig:pipeline}(c))
\begin{equation}
\mathcal{N}_{k+\epsilon}(\bm{z}) = 
\{ \bm{z}' \; | \;
\bm{z}'^\top\bm{z} > \epsilon \;\; \& \;\; {\text{among top-}k} \}, 
\label{eq:e-NN}
\end{equation}
where $\epsilon$ is the neighbourhood boundary threshold.
Adding such a similarity score constraint,
we aim to filter out the noisy tracklet pairs associated
by $k$-NN with low pairwise proximity.
The resulting neighbourhood sizes vary
from 0 to $k$ in accordance with
how many similar tracklets exist,
i.e. tracklet data adaptive.
This property is critical for model learning on unconstrained tracklets
without guarantee of a fixed number of reliable 
good-quality true matches.

After obtaining possibly matching tracklets
$\mathcal{N}_{k+\epsilon}(\bm{z}_i)$ for a specific tracklet $\bm{z}_i$,
we can compute the tracklet similarity weight as
their $L_1$ normalised quantity:
\begin{equation}
w(i,j) = \left\{
\begin{array}{ll}
\frac{\bm{z}_j^\top\bm{z}_i}
{\sum_{\bm{z}_k'\in\mathcal{N}_{k+\epsilon}} (z_k'^\top\bm{z}_i)},
& \text{if} \;\; \bm{z}_j \in \mathcal{N}_{k+\epsilon}(\bm{z}_i)\\
0, & \text{otherwise}
\end{array}
\right.
\label{eq:matching_weight}
\end{equation}
As such, only visually similar tracklets potentially
with minimal noisy image frames are encouraged (Eq. \eqref{eq:soft_TA_loss})
to be positive matches
in model learning.

\vspace{0.1cm}
\noindent{\bf Discussion.}
In formulation, our image-to-tracklet selective matching loss 
is similar as the 
instance loss \cite{wu2018unsupervised}.
Both are non-parametric variants of CE.
However, there are a few fundamental differences:
(1) The instance loss treats each individual image
as a class, whilst our loss considers tracklet-wise class. 
Conceptually, this introduces a two-tier hierarchical structure 
into the instance loss: local image and global tracklet.
(2) The instance loss does not consider the camera view structure.
In contrast, we uniquely combine the multi-task inference idea with 
tracklet classes for additionally exploiting the underlying correlation between
per-camera tracklet groups.
Moreover, our loss design
shares some spirit with the focal loss \cite{lin2017focal} both
using a modulating parameter for controlling the target degree
(noise measure in ours and imbalance measure in focal loss).
But they have more fundamental differences in addition to different formulations:
(1) The focal loss is parametric and supervised  vs. our non-parametric and unsupervised loss.
(2) The focal loss aims to solve the class imbalance 
between positive and negative samples in supervised learning,
whilst ours is for selective and robust image-to-tracklet matching
in unsupervised learning.

\subsection{Cross-Camera View Learning}
\label{sec:method_cross_cam_learning}

Besides per-camera view learning,
it is crucial to simultaneously consider cross-camera tracklet learning
\cite{LiTPAMI2019,li2018unsupervised}.
To this end, we need to similarly perform tracklet association
across camera views. 
%
We consistently utilise $k$-NN+$\epsilon$-NN for tracklet association
across different camera views. Specifically,
for a tracklet $\bm{z}$ we search
the nearest tracklets from different cameras (Fig. \ref{fig:pipeline}(c)):
\begin{equation}
\tilde{\mathcal{N}}_{k+\epsilon}(\bm{z}) = 
\{ \bm{z}' \; | \;
\bm{z}'^\top\bm{z} > \epsilon \;\; \& \;\; {\text{among top-}k} \}.
\label{eq:e-NN_cross_cam}
\end{equation} 
With the self-discovered 
cross-camera tracklet association ${\tilde{\mathcal{N}}_{k+\epsilon}(\bm{x}_i)}$
of a specific tracklet $\bm{z}_i$ 
which contains a training image frame $\bm{x}_i$, 
we then enforce a cross-camera image-to-tracklet matching loss function 
(Fig. \ref{fig:pipeline}(e)):
\begin{equation}\label{eq:CCM_loss}
\mathcal{L}_\text{ccm} =\sum_{\bm{z}' \in \tilde{\mathcal{N}}_{k+\epsilon}(\bm{z}_i)}
1-\bm{z}'^\top \bm{x}_i. 
\end{equation}
This loss encourages the image $\bm{x}_i$ to have as similar feature representation
of visually alike tracklets as possible in a cross-camera sense.
In doing so, person appearance variation across camera views 
is minimised if the image-to-tracklet association is correct.

\begin{algorithm}
	\caption{The STL model training procedure.} 
	\label{Algorithm}
	\textbf{Input:} Automatically generated raw tracklet data.\\
	\textbf{Output:} An optimised person re-id model. \\ [0.1cm]
	\textbf{for} $e=1$ \textbf{to}  \textsl{max\_epoch} \textbf{do} \\ 
	\hphantom{~~~~} 
	\textbf{if} $e$ in the first training stage epochs\\
	\hphantom{~~~~~~~~} 
	Update per-camera tracklet neighbourhood (Eq. \eqref{eq:e-NN}) \\
	\hphantom{~~~~~~~~} 
	\textbf{for} $t=1$ \textbf{to}  \textsl{per-epoch iteration number} \textbf{do} \\
	\hphantom{~~~~~~~~~~~~} 
	Feedforward a mini-batch of tracklet frame images \\
	\hphantom{~~~~~~~~~~~~}  
	Update the tracklet representations (Eq. \eqref{eq:tracklet_feat}) \\
	\hphantom{~~~~~~~~~~~~} 
	Compute per-camera matching loss (Eq. \eqref{eq:soft_TA_loss}) \\
	\hphantom{~~~~~~~~~~~~} 
	Update the model by back-propagation \\
	\hphantom{~~~~~~~~} 
	\textbf{end for} \\ [0.1cm]
	\hphantom{~~~~} 
	\textbf{else} 	/* The second training stage epochs */ \\
	\hphantom{~~~~~~~~}
	Update per-camera tracklet neighbourhood (Eq. \eqref{eq:e-NN}) \\
	\hphantom{~~~~~~~~}
	Update cross-camera tracklet neighbourhood (Eq. \eqref{eq:e-NN_cross_cam}) \\
	\hphantom{~~~~~~~~} 
	\textbf{for} $t=1$ \textbf{to}  \textsl{per-epoch iteration number} \textbf{do} \\
	\hphantom{~~~~~~~~~~~~} 
	Feedforward a mini-batch of tracklet frame images \\
	\hphantom{~~~~~~~~~~~~} 
	Update tracklet representations (Eq. \eqref{eq:tracklet_feat}) \\
	\hphantom{~~~~~~~~~~~~} 
	Compute STL model training loss (Eq. \eqref{eq:stul_loss}) \\
	\hphantom{~~~~~~~~~~~~}
	Update the network model by back-propagation \\
	\hphantom{~~~~~~~~} 
	\textbf{end for} \\ [0.1cm]
	\hphantom{~~~~} 
	\textbf{end if} \\
	\textbf{end for} 
\end{algorithm}

\begin{table*} 
	\begin{center}
	\setlength{\tabcolsep}{0.55cm}
	\begin{tabular}{|l||c|c|c|c|}
		\hline 
		\multirow{2}{*}{Dataset}
		& \multicolumn{2}{c|}{Training data}  & \multicolumn{2}{c|}{Test data}\\
		\cline{2-5}
		& \# Identity 	& \# Tracklet	& \# Identity 	& \# Tracklet \\
		\cline{1-5}
		MARS* \cite{zheng2016mars}	
		& 625 & 8,298 & 636 & 11,310 
		\\ \hline
		DukeMTMC-SI-Tracklet* \cite{ristani2016performance,LiTPAMI2019} 
		& 702 & 5,803 & 1,086 & 6,844 
		\\ \hline
		DukeMTMC-Raw-Tracklet* ({\bf New})
		& 702 & 7,427 & 1,105 & 8,950 
		\\ \hline \hline
		DukeMTMC-Raw-Tracklet ({\bf New})
		& 702 + unknown & 12,452 & 1,105 & 8,950 
		\\ \hline
	\end{tabular}
	\end{center}
	\caption{
		Dataset statistics and benchmarking setting.
		*: With tracklet selection.
	}
	\label{tab:dataset_stats}
\end{table*}

\subsection{Model Training}


\noindent {\bf Overall objective loss.}
By combining the per-camera and cross-camera learning constraints, 
we obtain the final model objective loss function as: 
\begin{equation}\label{eq:stul_loss}
\mathcal{L}_\text{STL}= 
{L}_\text{pcm} + \lambda \mathcal{L}_\text{ccm}
\end{equation}
where $\lambda$ is a balancing weight.
Trained jointly by $\mathcal{L}_\text{pcm}$ and $\mathcal{L}_\text{ccm}$,
the STL model is able to mine the discriminative re-id information
both within- and cross-camera views concurrently.
This overall loss function is differentiable therefore
allowing for end-to-end model optimisation.
For more accurate tracklet association between camera views,
we start to apply the cross-camera matching loss (Eq. \eqref{eq:CCM_loss})
in the middle of training as \cite{LiTPAMI2019}.
The STL model training process is summarised in Algorithm \ref{Algorithm}.



\vspace{0.1cm}
\noindent {\bf Tracklet representation.}
In our model formulation, we need to represent each tracklet as a whole.
To obtain this feature representation, we adopt a moving average strategy
\cite{lucas1990exponentially}
for computational scalability.
Specifically, we maintain a representation memory for each tracklet during training.
In each training iteration, given an input image frame $\bm{x}$,
we update its corresponding tracklet's feature vector $\bm{z}$ as:
\begin{equation}
\bm{z}=\frac{1}{2}(\bm{z}+\bm{x})
\label{eq:tracklet_feat}
\end{equation}
This scheme updates only the tracklets whose images are sampled 
by the current mini-batch in each iteration. 
Although not all the tracklets are updated and synchronised 
along with the model training, 
the discrepancy from their accurate representations is 
supposed to be marginal due to the small model learning speed
therefore matters little.

\vspace{0.1cm}
\noindent{\bf Model parameter setting.}
In unsupervised re-id learning,
we have {\em no} access to labelled training data for 
model parameter selection by cross-validation.
Also, it is improper to use any test data for model parameter tuning
which is not available in real-world application.
All the parameters of a model are usually estimated empirically.
Moreover, the identical parameter setting should be used
for all the different datasets in domain scalability
and generic large scale application considerations.
With this principle, we set the parameters of our STL model for all 
different tests and experiments as:
$\tau=0.1$ for Eq. \eqref{eq:prob},
$\lambda=10$ for Eq. \eqref{eq:stul_loss},
$\epsilon=0.7$ and $k=1$ for Eq. \eqref{eq:e-NN} and \eqref{eq:e-NN_cross_cam}.
Otherwise parameter settings may give better model performance
on specific tests. 
But they are not exhaustively considered in our study, because
this often assumes extra domain knowledge which
is {\em not} generally available 
therefore making the performance evaluation less realistic
and less generic. 

To minimise the negative effect of inaccurate cross-camera, STL begins the model training with per-camera image-to-tracklet selective matching loss Eq. \eqref{eq:soft_TA_loss} during the first training stage, and then add cross-camera image-to-tracklet selective matching loss Eq. \eqref{eq:CCM_loss} in the second training stage. We do not incorporate the ccm loss until in the second training stage due to insufficiently reliable feature representations for cross-camera tracklet matching in the beginning of training stage.


\section{Experiments}

\subsection{Experimental Setup}
\label{sec:exp_setup}

\noindent \text{\bf Datasets.}
To evaluate the proposed STL model, we tested on two 
publicly available person tracklet datasets:
{\em MARS} \cite{zheng2016mars}, 
and {\em DukeMTMC-SI-Tracklet} \cite{zheng2016mars,LiTPAMI2019}.
The dataset statistics and test settings are given 
in Table \ref{tab:dataset_stats}.
The previous two datasets both contain
only {\em manually selected} person tracklet data 
therefore presenting {\em less realistic} unsupervised
learning scenarios.
To enable more realistic algorithm test, 
we introduced a new {\em raw} tracklet person re-id dataset.

As DukeMTMC-SI-Tracklet,
we used the DukeMTMC tracking videos \cite{ristani2016performance}.
To extract person tracklets,
we leveraged an efficient detector-and-tracker model \cite{girdhar2018detect}.
and a graph association method. 
%
From all the DukeMTMC videos captured from 8 distributed surveillance cameras, 
we obtained 21,402 person tracklets, including a total of 1,341,096 
bounding box images.
Detection and tracking errors are inevitable as shown in Fig. \ref{fig:raw_tkls}(b).
In reality, we usually assume no manual efforts for cleaning
tracklet data but expect the unsupervised learning algorithm to be sufficiently
robust to any errors and noise.
We therefore keep all tracklets.
In spirit of DukeMTMC-SI-Tracklet, we call the newly introduced
dataset {\em DukeMTMC-Raw-Tracklet}.

For the DukeMTMC-Raw-Tracklet dataset benchmarking,
we utilised a similar method as \cite{LiTPAMI2019}
to automatically label the identity classes of test person tracklets.
This is for enabling model performance test.
We used the same 1,105 test person identity classes
as DukeMTMC-SI-Tracklet for allowing an apple-to-apple comparison 
between datasets.
The number of training person identity classes
is unknown due to no manual annotation, a natural property of 
unconstrained raw tracklets
in real-world application scenarios.

\begin{figure}[t] \label{fig:dataset_tkl}
	\centering
	\includegraphics[width=0.45\textwidth]{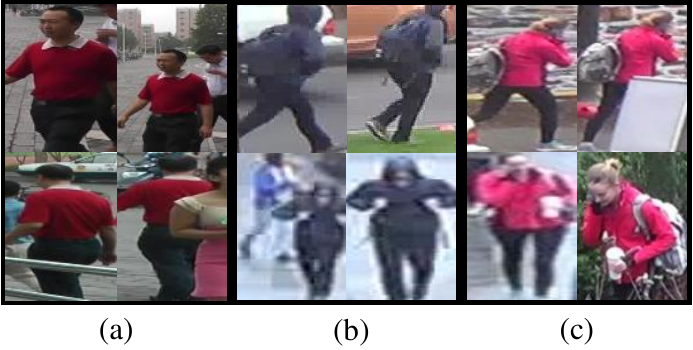}
	\caption{Cross-camera tracklet matching pairs from 
		(a) MARS, 
		(b) DukeMTMC-SI-Tracklet, and  
		(c) DukeMTMC-Raw-Tracklet. 
	}
\end{figure}

To assess the exact effect of manual selection to
unsupervised model learning,
we further selected out the tracklets of the same 702 training person identities
as {\em DukeMTMC-SI-Tracklet}.
Along with the same test data,
we built another version of dataset
with selected training tracklets as
the conventional datasets \cite{zheng2016mars,LiTPAMI2019}.
We name this dataset {\em DukeMTMC-Raw-Tracklet*}.
The statistics of both datasets with and without selection are described in
Table \ref{tab:dataset_stats}.

\vspace{0.1cm}
\noindent \textbf{Performance metrics.}
For model performance measurement, we used the Cumulative Matching Characteristic (CMC) and mean Average Precision (mAP) metrics. 

\vspace{0.1cm}
\noindent \textbf{Implementation details.}
We used an ImageNet pre-trained ResNet-50 \cite{he2016deep} as the backbone
for our STL model.
The re-id feature representation is 
$L_2$ normalised with 128 dimensions each. 
Person bounding box images were resized to $256\!\times\!128$. 
%
%
We adopted the Stochastic Gradient Descend (SGD) optimiser.
We set
the learning rate is $3 \!\times\! 10^{-5}$,
the mini-batch size to 384,
and the epoch number to 20.
The first training stage begins from the first epoch, 
and the second training stage from the tenth epoch.


\begin{table*}[h] 
	\begin{center}
	\setlength{\tabcolsep}{0.05cm}
	\begin{tabular}
		{|l||c|c|c|c||c|c|c|c||c|c|c|c||c|c|c|c|}
		\hline
		\multirow{2}{*}{\bf Methods}
		& \multicolumn{4}{c||}{MARS*}
		& \multicolumn{4}{c||}{Duke-SI-TKL*}
		& \multicolumn{4}{c||}{Duke-Raw-TKL*}
		& \multicolumn{4}{c|}{Duke-Raw-TKL } \\
		\cline{2-17}
		& R1 	& R5	& R20 	& mAP
		& R1 	& R5	& R20 	& mAP
		& R1 	& R5	& R20 	& mAP
		& R1 	& R5	& R20 	& mAP \\ 
		\hline \hline
		GRDL \cite{kodirov2016person}
		& 19.3	& 33.2	& 46.5	& 9.6
		& -	& -	& -	& - 	& -	& -	& -	& - 	& -	& -	& -	& -	\\ 
		UnKISS \cite{khan2016unsupervised}				
		& 22.3	& 37.4	& 53.6	& 10.6
		& -	& -	& -	& - 	& -	& -	& -	& -  	& - & -	& -	& - \\
		\hline
		SMP$^\dagger$ \cite{liu2017stepwise}
		& 23.9   	& 35.8		& 44.9		& 10.5 			
		& -	& -	& -	& - 	& -	& -	& -	& -  	& - & -	& -	& - \\
		DGM+IDE$^\dagger$ \cite{ye2017dynamic}
		& 36.8  	& 54.0  	& 68.5 		& 21.3
		& -	& -	& -	& - 	& -	& -	& -	& -  	& - & -	& -	& - \\
		RACE$^\dagger$ \cite{ye2018robust}
		& 43.2  	& 57.1 	 	& 67.6 		& 24.5
		& -	& -	& -	& - 	& -	& -	& -	& -  	& - & -	& -	& -
		\\
		\hline
		TAUDL \cite{li2018unsupervised}
		& 43.8 		& 59.9 		& 66.0 		& 29.1
		& 26.1		& 42.0		& 57.2		& 20.8 
		& -	& -	& -	& -
		& -	& -	& -	& -
		\\
		DAL \cite{chen2018deep}
		& 46.8  	& 63.9  	& 71.6 		& 21.4
		& -	& -	& -	& - 	& -	& -	& -	& -  	& - & -	& -	& -
		\\
		BUC \cite{lin2019aBottom}
		& 51.1		& 64.2 		& 72.9 		& 26.4
		& 30.6		& 43.9		& 51.7		& 16.7 
		& 38.6		& 50.1		& 61.9		& 20.1 
		& 31.1		& 41.3		& 52.0		& 15.1
		\\
		UTAL \cite{LiTPAMI2019}
		& 49.9 		& 66.4 		& 77.8 		& 35.2
		& 43.8		& 62.8		& 76.5		& 36.6 
		& 48.7		& 62.9		& 76.6		& 38.4 
		& 41.3  	& 55.7		& 71.3		& 31.8 \\
		\hline
		\bf {STL (Ours)}
		& \bf{54.5}	& \bf{71.5}	 & \bf{82.0} 	& \bf{37.2}
		& \bf{46.7}	& \bf{65.4}	 & \bf{78.1}	& \bf{38.9}
		& \bf{55.4}	& \bf{71.7}	 & \bf{79.0}	& \bf{41.5}
		& \bf{56.1}	& \bf{69.6}	 & \bf{80.9} 	& \bf{41.5}
		\\
		\hline
	\end{tabular}
	\end{center}
	\caption{Unsupervised person re-id performance on tracklet benchmarking datasets.
		$^\dagger$: Use one-shot labels. 
		*: With manual tracklet selection. TKL = Tracklet.}
	\label{tab:vid_SOTA}
\end{table*}

\subsection{Comparisons to the State-Of-The-Art Methods}
\label{sec:exp_SOTA}


\noindent {\bf Competitors.}
We compared the proposed STL method 
with three different modelling strategies:
(a) 
{\em Hand-crafted feature based methods} 
(GRDL \cite{kodirov2016person},
UnKISS \cite{khan2016unsupervised}),
(b) 
{\em One-shot learning methods}
(SMP \cite{liu2017stepwise},
DGM+IDE \cite{ye2017dynamic},
RACE \cite{ye2018robust}),
(c) 
{\em Unsupervised deep learning models}
(TAUDL \cite{li2018unsupervised},
DAL \cite{chen2018deep},
BUC \cite{lin2019aBottom},
UTAL \cite{LiTPAMI2019}).


\vspace{0.1cm}
\noindent {\bf Results.}
Table \ref{tab:vid_SOTA} compares the re-id performance.
We have the following main observations:
\\
{\bf (1)} 
Hand-crafted feature based methods (GRDL and UnKISS)
produce the weakest performance.
This is due to the poor discriminative ability
of manually designed features
and the lacking of end-to-end model optimisation.
\\
{\bf (2)} 
One-shot learning methods (SMP, DGM, RACE) improve the re-id model
generalisation capability.
However, their assumption on one-shot training data
would limit their application scalability due to the 
need for some amount of person identity labelling per domain.
\\
{\bf (3)} 
The more recent unsupervised deep learning methods
(TAUDL, DAL, BUC\footnote{We utilised the officially released codes of BUC \cite{lin2019aBottom} with the default parameter settings, 
and we used the single parameter setting for BUC on all the tests. 
In contrast, the authors of BUC seemingly used the labelled test data to 
tune the model parameters which is {\em improper} for unsupervised learning.
As a result, the results on MARS were reported differently.},
UTAL) further push the boundary of model performance.
However, all these existing methods are outperformed clearly by the proposed
STL model, in particularly on the unconstrained raw tracklets training data.
This suggests the overall performance advantages of our model
over the strong alternative methods.
\\
{\bf (4)}
Existing methods BUC and UTAL both suffer from the 
noisy data in unconstrained raw tracklets, as indicated by
their significant performance drop from DukeMTMC-Raw-Tracklet* (with tracklet selection) 
to DukeMTMC-Raw-Tracklet.
This suggests that the data selection
in dataset benchmarking simplifies the model learning task,
i.e. less challenging than the realistic setting.
The proposed DukeMTMC-Raw-Tracklet dataset is designed particularly
for addressing this problem.
\\
{\bf (5)}
Our STL model is shown to be more robust against more noisy tracklet data,
with little re-id performance changes. 
This suggests the superiority of our image-to-tracklet selective matching
in dealing with more noisy unconstrained tracklets. 
\\
{\bf (6)}
While both UTAL and STL adopt multi-task learning model design,
STL is also superior on all three datasets with manual selection. 
This suggests that assuming a fixed number of true matches (due to using $k$-NN) is 
suboptimal even for carefully constructed training data,
and the modelling superiority of our image-to-tracklet selective matching
in handling unconstrained raw tracklet data with more noise.


\subsection{Further Analysis and Discussions}
\label{sec:exp_component}

To provide more insight and interpretation into the performance advantages of
our STL method, we analysed key model designs 
on two large tracklet re-id datasets:
MARS and DukeMTMC-Raw-Tracklet.

\vspace{0.1cm}
\noindent {\bf Loss design for per-camera view learning. }
The loss function for per-camera view learning is a key component in STL.
We compared our image-to-tracklet selective matching (ITSM) loss 
(Eq. \eqref{eq:soft_TA_loss}) 
with the conventional cross-entropy (CE) loss (Eq. \eqref{eq:CE_loss}.)
Table \ref{tab:per_cam_loss_eval} shows that
the ITSM loss is significantly more effective 
especially on DukeMTMC-Raw-Tracklet dataset.
This suggests the superiority of our loss design
in handling noisy tracklets, thanks to
its data adaptive and selective learning capability.

\begin{table}[h] 
	\begin{center}
	\setlength{\tabcolsep}{0.28cm}
	\begin{tabular}
		{|c||c|c||c|c|}
		\hline
		Dataset		
		& \multicolumn{2}{c||}{MARS*}
		& \multicolumn{2}{c|}{Duke-Raw-TKL}
		\\ \hline
		Loss  & Rank-1 & mAP	& Rank-1 & mAP \\ 
		\hline \hline
		CE & 43.8 & 31.4 & 22.4 & 15.3
		\\
		\hline
		\bf ITSM
		&\bf 48.2 &\bf 32.2 &\bf 49.1 &\bf 34.2 \\
		\hline
	\end{tabular}
	\end{center}
	\caption{Evaluate loss design for per-camera view learning.
		ITSM: per-camera Image-to-Tracklet Selective Matching.}
	\label{tab:per_cam_loss_eval}
\end{table}

\vspace{-0.1cm}
\noindent {\bf Image-to-tracklet selective effect.}
We tested the data selective 
effect by controlling the temperature parameter
$\tau$ in Eq. \eqref{eq:prob}.
It is one of the key factors for our method
to be able to select possibly well matching tracklets
and deselect potentially noisy tracklets.
Table \ref{tab:temp} shows several key observations.
(1) With $\tau\!=\!1$ we impose no selection effect in matching,
the model generalisation performance degrades significantly.
(2) When setting small values ($<$0.2) to $\tau$, 
as expected the model performance is dramatically boosted.
This is due to the modulating effect of our loss function to
the selective matching between images and tracklets.
It is also observed that more gain is obtained in case of unconstrained raw tracklets
due to more noise and distraction.
(3) The optimal value is around $\tau\!=\!0.1$ consistently,
suggesting the generic benefit of a single setting. 

\begin{table}[h] 
	\begin{center}
	\setlength{\tabcolsep}{0.28cm}
	\begin{tabular}
		{|c||c|c||c|c|}
		\hline
		Dataset
		& \multicolumn{2}{c||}{MARS*}
		& \multicolumn{2}{c|}{Duke-Raw-TKL} \\
		\hline
		$\tau$ & Rank-1 & mAP
		& Rank-1 & mAP \\ 
		\hline \hline
		1
		& 15.7 & 9.8  &10.4 & 6.3\\
		\hline
		0.5
		& 25.3 & 15.1  &14.6 &11.4\\
		\hline
		0.2
		& 44.7 &29.9	& 41.7 & 30.5 \\
		\hline
		0.1
		&\bf 54.5 &\bf37.2  &\bf56.1 &\bf41.5\\
		\hline
		0.05
		& 47.6 & 32.5  &47.7 &35.7\\
		\hline
	\end{tabular}
	\end{center}
	\caption{Evaluate the temperature $\tau$.
	}
	\label{tab:temp}
\end{table}

\noindent {\bf Benefit of cross-camera view learning.}
We evaluated the efficacy of.
cross-camera view learning
Table \ref{tab:eff_cross_cam_loss} shows that
significant re-id accuracy gains can be obtained.
This verifies the cross-camera image-to-tracklet matching loss (Eq. \eqref{eq:CCM_loss}) on top
of per-camera view learning
(Eq. \eqref{eq:soft_TA_loss}).

\begin{table}[h] 
	\begin{center}
	\setlength{\tabcolsep}{0.2cm}
	\begin{tabular}
		{|c||c|c||c|c|}
		\hline
		Dataset		
		& \multicolumn{2}{c||}{MARS*}
		& \multicolumn{2}{c|}{Duke-Raw-TKL}
		\\ \hline
		CCM  & Rank-1 & mAP	& Rank-1 & mAP \\ 
		\hline \hline
		\xmark
		& 48.2	& 32.2		& 49.1	& 34.2 \\
		\hline
		\checkmark
		& \bf{54.5} 	& \bf{37.2}		& \bf{56.1} 	& \bf{41.5} \\
		\hline
	\end{tabular}
	\end{center}
	\caption{Effect of cross-camera matching (CCM) learning.}
	\label{tab:eff_cross_cam_loss}
\end{table}

\begin{figure}[h] 
	\centering
	\subfigure[]{\includegraphics[width=0.48\linewidth]{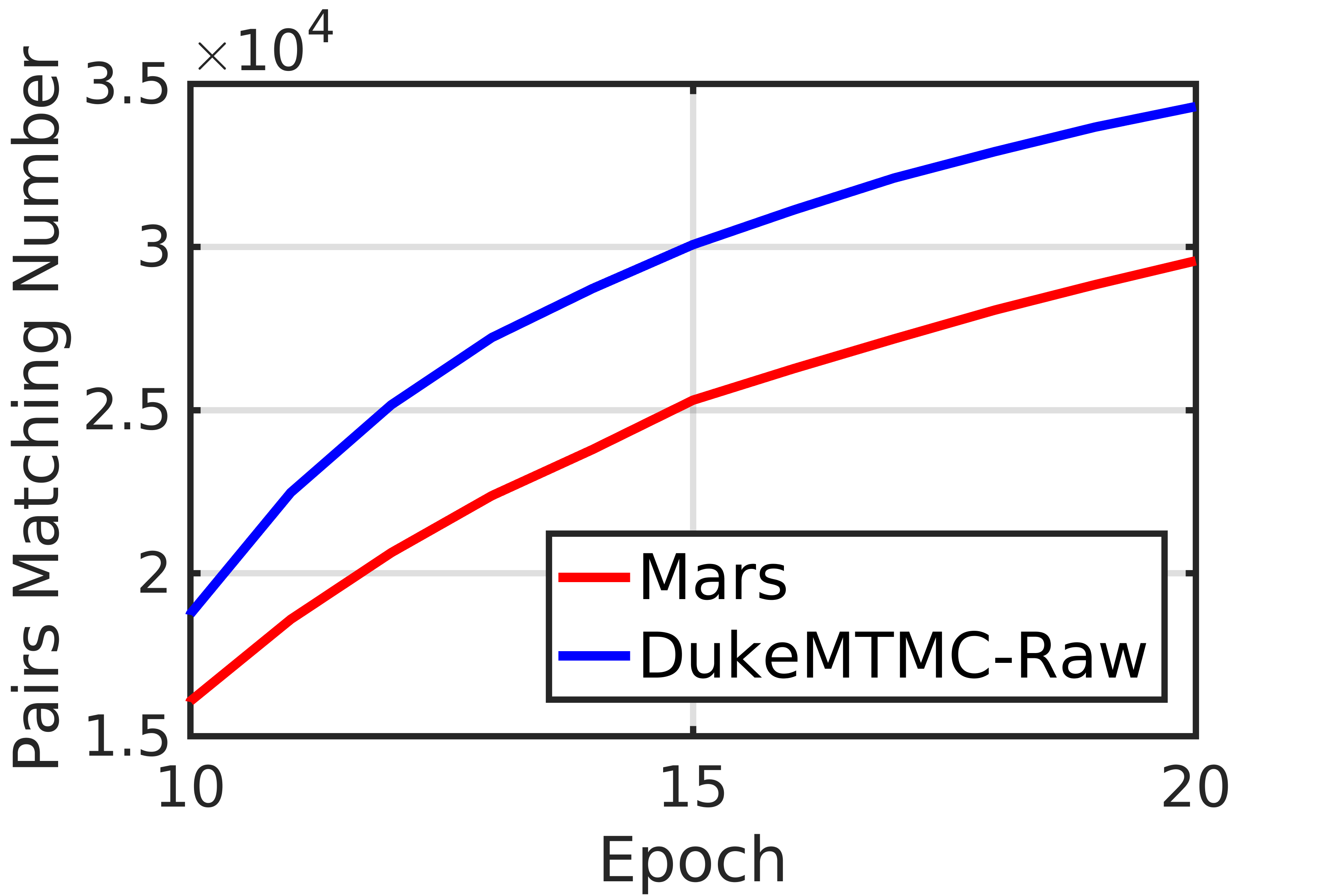}}
	\subfigure[]{\includegraphics[width=0.48\linewidth]{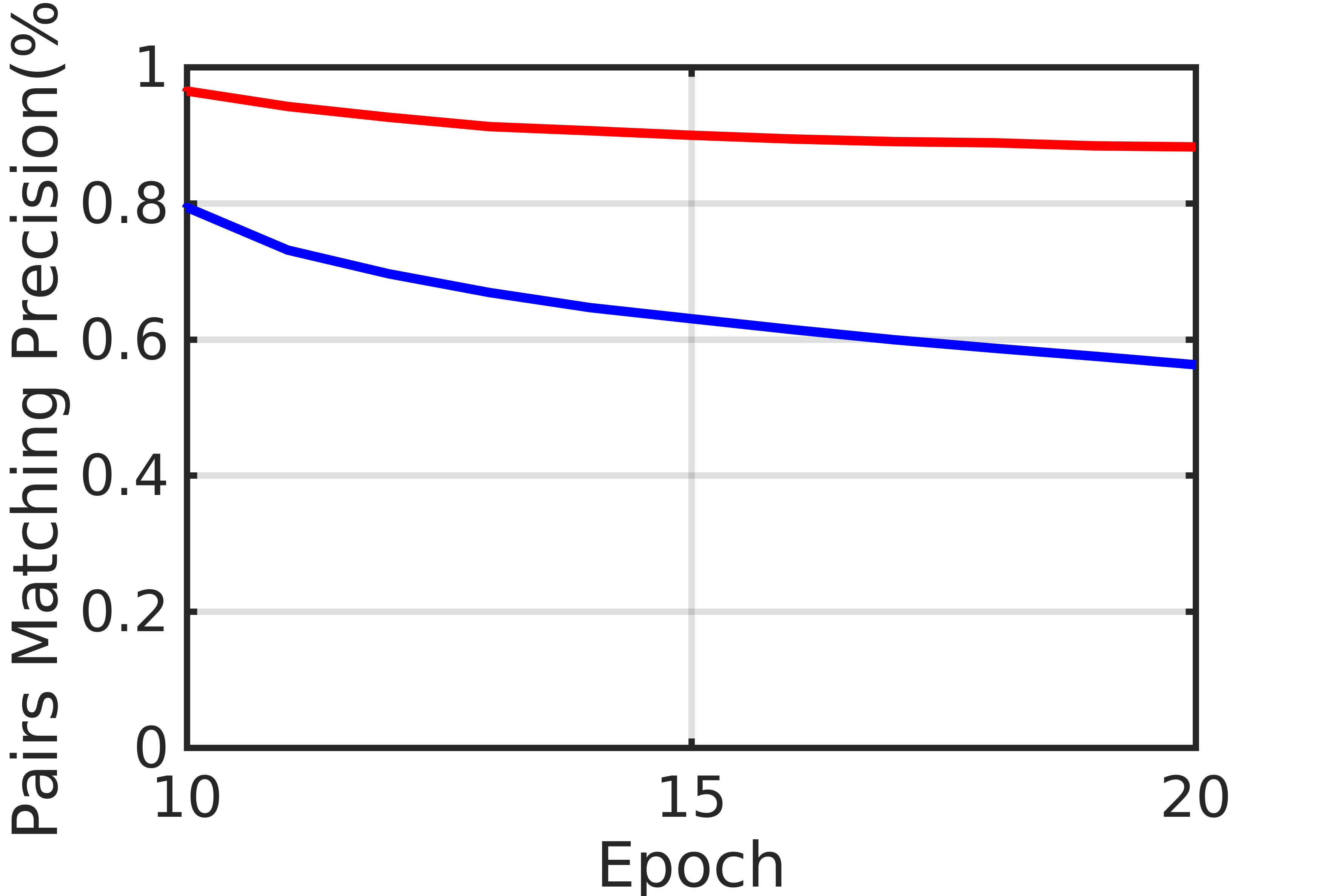}}
	\caption{The (a) number and (b) precision of cross-camera tracklet pairs
		discovered during training.  
	}
\label{fig:cross_cam_rate}
\end{figure}

To examine further cross-camera matching,
we tracked the number and accuracy of tracklet association in training.
Figure \ref{fig:cross_cam_rate} shows that
the number of cross-camera tracklet pairs grow dramatically
whilst the matching accuracy drops slightly or moderately 
along the training.
This justifies the positive effect of 
cross-camera tracklet association.
That is, most pairs are correct true matches, 
providing model training with discriminative information for person appearance variation
across camera views.
Besides, we also observe further room for 
more accurate tracklet association.

\noindent {\bf Tracklet association strategy. }
We tested the effect of $\epsilon$-NN on top of 
$k$-NN (Eq. \eqref{eq:e-NN} and \eqref{eq:e-NN_cross_cam}) as
the tracklet association strategy in STL.
We set $k=1$ and $\epsilon=0.7$, the default setting of our model.
Table \ref{tab:eff_K} shows that
using $k$-NN only is inferior to  $k$-NN+$\epsilon$-NN.
This suggests the data adaptive benefit of $\epsilon$-NN
particularly in handling unconstrained raw tracklet data,
verifying our design of additionally leveraging $\epsilon$-NN
for tracklet association except $k$-NN.

\begin{table}[h] 
	\begin{center}
	\setlength{\tabcolsep}{0.1cm}
	\begin{tabular}
		{|l||c|c||c|c|}
		\hline
		\multirow{2}{*}{}
		Dataset & \multicolumn{2}{c||}{MARS*}
		& \multicolumn{2}{c|}{Duke-Raw-TKL} \\
		\hline 
		Strategy& Rank-1 & mAP
		& Rank-1 & mAP \\ 
		\hline \hline
		$k$-NN 
		& 51.0		& 32.7
		& 51.0		& 37.4 \\
		\hline 
		$k$-NN+$\epsilon$-NN
		& \bf{54.5} 	& \bf{37.2}
		& \bf{56.1} 	& \bf{41.5} \\
		\hline
	\end{tabular}
	\end{center}
	\caption{$k$-NN versus $\epsilon$-NN in tracklet association.
}
	\label{tab:eff_K}
\end{table}

\noindent {\bf Sensitivity of tracklet association threshold.}
We tested the model performance sensitivity of setting 
the tracklet matching similarity threshold $\epsilon$
(Eq. \eqref{eq:e-NN} and \eqref{eq:e-NN_cross_cam}).
Table \ref{tab:eff_thres} shows that
re-id accuracies vary with the change of $\epsilon$ as expected.
This is because $\epsilon$ controls what image-to-tracklet matching pairs 
are used in objective loss functions during training.
Importantly, $\epsilon$ is not very sensitive 
with a good value range (around $0.7\!\sim\!0.8$) giving strong model performance.
This robustness is a critical property of our method, 
since when applied to diverse tracklets data with unconstrained conditions, 
label supervision is not available for hyper-parameter cross-validation.

\begin{table}[h] 
	\begin{center}
	\setlength{\tabcolsep}{0.1cm}
	\begin{tabular}
		{|c||c|c||c|c|}
		\hline
		\multirow{2}{*}{}
		Dataset & \multicolumn{2}{c||}{MARS*}
		& \multicolumn{2}{c|}{Duke-Raw-TKL} \\
		\hline 
		$\epsilon$& Rank-1 & mAP
		& Rank-1 & mAP \\ 
		\hline \hline
		0.9
		& 48.7 & 30.4  & 48.3 & 33.3 \\ \hline
		0.8
		& 54.2 & 36.2 & 55.3 & 41.3  \\ \hline
		0.7 
		& \bf 54.5 & \bf 37.2 &\bf 56.1 &\bf 41.5 \\ \hline
		0.6
		& 50.7 & 34.1 & 47.7 & 35.2 \\
		\hline 
	\end{tabular}
	\end{center}
	\caption{Evaluate the tracklet association threshold $\epsilon$.
		}
	\label{tab:eff_thres}
\end{table}


\section{Conclusion}

We presented a {\em selective tracklet learning} (STL) approach, 
which aims to address the limitations of
both the existing
supervised person re-id methods and unsupervised tracklet learning methods
concurrently.
Specifically, STL is able to learn 
discriminative and generalisable re-id model
from unlabelled raw tracklet datasets.
This eliminates the artificial assumptions
on exhaustive person ID labelling as by
supervised re-id methods,
and manual filtering as by existing tracklet unsupervised learning models.
We also introduced an unconstrained raw tracklet person re-id benchmark, DukeMTMC-Raw-Tracklet.
Extensive experiments 
show the superiority and robustness advantages 
of STL
over the state-of-the-art unsupervised learning re-id methods 
on three tracklet person re-id benchmarks.



{\small
\bibliographystyle{ieee_fullname}
\bibliography{iccv19}
}

\end{document}